\pdfoutput=1

\documentclass[]{article}
\usepackage{authblk}

\usepackage{framed,multirow}

\usepackage{amssymb}
\usepackage{latexsym}

\usepackage{subfigure} 
\usepackage{algorithm}
\usepackage{algorithmic}
\usepackage{multirow}
\usepackage{graphicx}

\usepackage{url}
\usepackage{xcolor}
\definecolor{newcolor}{rgb}{.8,.349,.1}

\newcommand*{\addheight}[2][.5ex]{%
	\raisebox{0pt}[\dimexpr\height+(#1)\relax]{#2}}%

\date{September 2018}

\title{Supervised Learning of the Next-Best-View for 3D Object Reconstruction\footnote{Submitted to Pattern Recognition Letters.}}

\author[1]{Miguel Mendoza} 
\author[1,2]{J. Irving Vasquez-Gomez}
\author[1]{Hind Taud}
\author[3]{L. Enrique Sucar}
\author[4]{Carolina Reta}

\affil[1]{Centro de Innovaci\'on y Desarrollo Tecnol\'ogico en C\'omputo, Instituto Polit\'ecnico Nacional, Ciudad de M\'exico, M\'exico.}
\affil[2]{Consejo Nacional de Ciencia y Tecnolog\'ia (CONACYT), Ciudad de M\'exico, M\'exico.}
\affil[3]{Instituto Nacional de Astrof\'isica \'Optica y Electr\'onica, Puebla, M\'exico.}
\affil[4]{Department of IT, Control, and Electronics, CONACYT-CIATEQ A. C., Av. Diesel Nacional \#1 Ciudad Sahag\'un,  Hidalgo 43990, M\'exico.}

\begin{document}

\maketitle

\begin{abstract}
Motivated by the advances in 3D sensing technology and the spreading of low-cost robotic platforms, 3D object reconstruction has become a common task in many areas. Nevertheless, the selection of the optimal sensor pose that maximizes the reconstructed surface is a problem that remains open. \textcolor{black}{It} is known in the literature as the next-best-view \textcolor{black}{planning problem}. In this paper, we propose a novel next-best-view planning scheme based on supervised deep learning. The scheme contains an algorithm for automatic generation of datasets and an original three-dimensional convolutional neural network (3D-CNN) used to learn the next-best-view. Unlike previous work where the problem is addressed as a search, the trained 3D-CNN directly predicts the sensor pose. We present a comparison of the proposed network against a similar net, and we present several experiments of the reconstruction of unknown objects validating the effectiveness of the proposed scheme.
\end{abstract}

\section{Introduction}

Autonomous three-dimensional (3D) object reconstruction or inspection is a computer vision task with applications to many areas, for example, robotics or cultural heritage conservation. It consists of generating a 3D model from a physical object by sensing its surface from several points of views. When the object is unknown, the task is performed iteratively in four steps: sensor positioning, sensing, registration and planning of the next sensor location \cite{Scott_03}. In this work, we are interested in the challenging step of planning. According to the literature, the addressed challenge is called next-best-view (NBV) problem and it has been defined as the task of computing the sensor position and orientation that maximizes the object's surface \cite{connolly1985determination}.

So far, most of the state-of-the-art techniques for NBV planning have been manually designed depending on the representations, needs, and constraints of reconstruction scene. Some methods, called synthesis methods, rely on analyzing the current information about the surface and directly synthesize the NBV \cite{chen2005vision}, \cite{Maver_93}. Synthesis methods are fast but their performance is usually decreased by objects with self-occlusions \cite{Scott_03}. Another type of methods, called search-based, define a utility function and then perform a search over the feasible sensor positions in order to optimize the utility function. For instance, \cite{yamauchi1997frontier} and \cite{bircher2016receding} represent the information with a probabilistic uniform grid, and they define as relevant features the frontier cells, as a result, they perform a search of the sensor position and orientation from where the maximum number of frontier cells is observed. Recent utility functions are based on the information gain contained in the model, \cite{Isler16}. In such cases, the target is to find the sensor pose that observes the cells (voxels) whose entropy is higher. Search-based methods are time-consuming given that the utility function has to be evaluated multiple times. Therefore, there is still a need of methods with the following characteristics: i) effective in spite of the object's shape and ii) efficient in terms of computation time. 

On the other hand, leveraged by technological advances in parallel processing hardware, the machine learning technique called deep learning (DL) \cite{lecun2015deep} have dramatically improved the state-of-the-art in several areas, such as optical character recognition, two or three dimensional object recognition and classification \cite{krizhevsky2012imagenet}. One of the main advantages of DL is to automatically discover the features that describe an entity or situation. 

Our hypothesis is that the current NBV planning paradigm, usually seen as an optimization problem \cite{foissotte2009two}, can be modeled as a supervised learning problem. The trained function should take as input the partially reconstructed model and it should output the NBV. To the best of our knowledge, the use learning approaches, specifically deep learning, for NBV prediction is an unexplored path. Even though new approaches are trying to learn the utility function \cite{hepp2018arxiv}, the whole NBV prediction has not been addressed.

In this paper, we propose a supervised learning scheme to predict the NBV. We provide a methodology for generating datasets and an original 3D convolutional neural network. The scheme has been modeled as a classification problem where to a given partial model a possible sensor pose (class) is assigned. To validate the proposed scheme we have i) generated a dataset with more than 15,000 examples, ii) trained the proposed network, called NBV-Net, iii) compared its accuracy against a related network, and iv) tested the trained network in the reconstruction of several unknown objects. Our experiments have shown that NBV-Net improves the accuracy of related networks, and it has been capable of predicting the NBV during the reconstruction of several unknown objects (do not seen by the network during training) reaching a reconstruction coverage higher than 90 percent.

\section{Next-best-view learning problem}
\begin{figure}[tb]
\centering
\subfigure[Iteration one]{\includegraphics[width=0.23\textwidth]{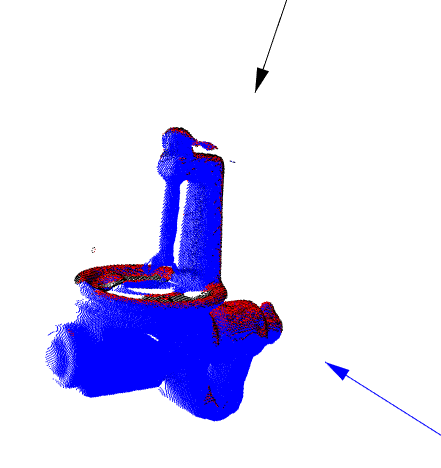}}
\subfigure[Iteration two]{\includegraphics[width=0.20\textwidth]{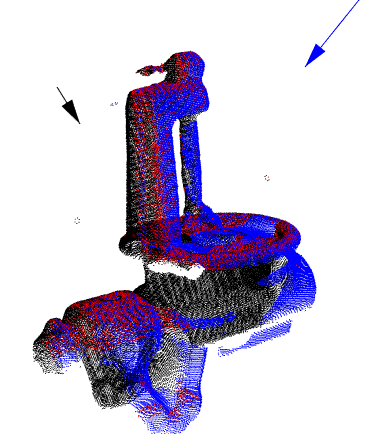}}
\subfigure[Iteration three]{\includegraphics[width=0.20\textwidth]{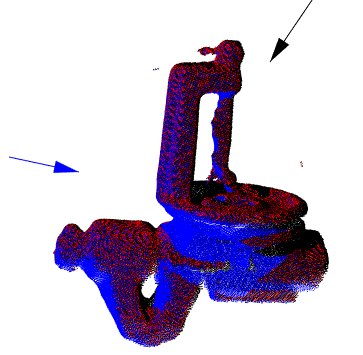}}
\subfigure[Iteration four]{\includegraphics[width=0.24\textwidth]{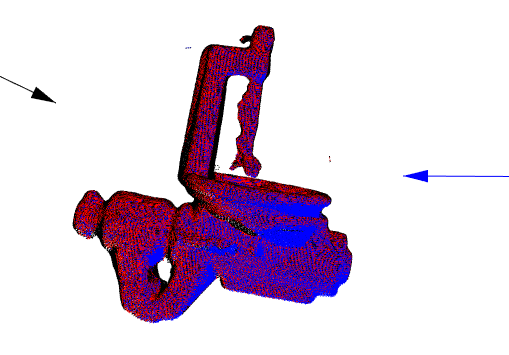}}
\caption{\textcolor{black}{Example of a 3D reconstruction. For each iteration, the current accumulated point cloud is displayed in black and the current sensor's pose is draw as a black arrow. Based on the current information, the NBV (blue arrow) is computed. The perception made at the NBV is drawn in blue. The overlap between the perception at the NBV and the accumulated point cloud is drawn in red. Finally, the new perception is registered to the accumulated point cloud. Note that in the next iteration the previous NBV is now the current sensor's pose. Figure best seen in color.}} \label{fig:pcl}
\end{figure}

The NBV concept originally rose up in the context of 3D object reconstruction \cite{connolly1985determination}. In such a task, a sensor is placed at several poses around the object of interest in order to get its 3D shape. Due to the limited information about the object shape, the reconstruction is done iteratively by the steps of positioning, perception, registration and planning the NBV. Fig. \ref{fig:pcl} shows four iterations of the reconstruction of an example object. During the positioning, the sensor is placed at a given pose. The sensor's pose (also named view) defines its position and orientation, namely $v = (x,y,z,\alpha,\beta,\gamma)^T$, where $\alpha$ is a rotation about $x$ axis, $\beta$ is a rotation about $y$ axis and $\gamma$ is a rotation about $z$ axis. At the perception step, the object's surface is measured and a point cloud ($z$) from the shape is obtained. After the perception, the acquired point cloud is registered to a single model \cite{icp}. As the reconstruction advances, the gathered information is stored into a partial model \cite{hornung13auro}. In this work, we use a uniform probabilistic occupancy grid, $M$, where each voxel has an associated probability that represents the likelihood that part or all the object's surface is inside the voxel's volume.

With respect to the NBV computation, so far, it has been treated as a search problem, where the target is to find the view that maximizes a metric. Unlike that approaches, in this work, we aim to directly predict the NBV based on the information provided by the partial model. Such prediction must be based on knowledge obtained from previous reconstructions. Henceforth, we define the NBV learning problem as the task of learning a function 
\begin{equation}
f(M):\mathbb{R}^{n} \rightarrow \mathbb{R}^{3} \times SO(3)
\label{eq:f_learning}
\end{equation} so that the perception at $v = f(M)$ increases the information contained in the partial model about the object. The input of $f$ is the voxel grid and it is written in eq. (\ref{eq:f_learning}) as $\mathbb{R}^{n}$ where $n$ is the amount of voxels in the grid. If we consider only the occupancy probability then the domain of $f$ is only \textcolor{black}{$[0,1]^n$}. The output is directly a sensor pose where a perception should be taken.

The proposed data-driven approach decomposes the NBV computation in two steps, the first step is the learning process of $f$, which has to be carried out off-line and using a plenty amount of examples from previous reconstructions; the second step is the prediction where $f$ is used to estimate the NBV. The latter step is executed on-line during the reconstruction and it is hoped to be as quick as possible.

In general, learning $f$ implies to solve a regression problem and several challenges need to be addressed, therefore our first approach is to face the problem as a classification problem.

\section{Dataset Generation}

Deep learning approaches require vast amounts of examples to reach successful performances. In this NBV-learning approach, we require a dataset with examples composed by an input (probabilistic grid) and a correct output (NBV).	One of the challenges is to provide the ``correct" or ``ground truth" NBV. It is an important challenge because it implies to solve the NBV, which remains as an NP-Hard problem even though we could have the complete information about the object. Therefore, in this section, we propose a methodology to obtain \textcolor{black}{a resolution-optimal} NBV, in other words, we perform an exhaustive search over a finite set of candidate views using the complete information about the object. Making use of human expertise to label the dataset has been discarded since there are inherent constraints that even for a human expert are not easy to estimate, one of them is the registration constraint which is needed to build a unified model.

The proposed methodology works as follows. Given a demo object (an object from which we known the whole surface), 1) we establish a discrete NBV search space and then 2) \textcolor{black}{we iteratively reconstruct it while we generate resolution-optimal NBVs. Below we describe the discrete search space, then we provide the definition and computation of a resolution-optimal NBV and finally, we provide the iterative process for building the dataset.}

\subsection{Search Space Generation}
\begin{figure}[tb]
\centering
\includegraphics[width=0.5\textwidth]{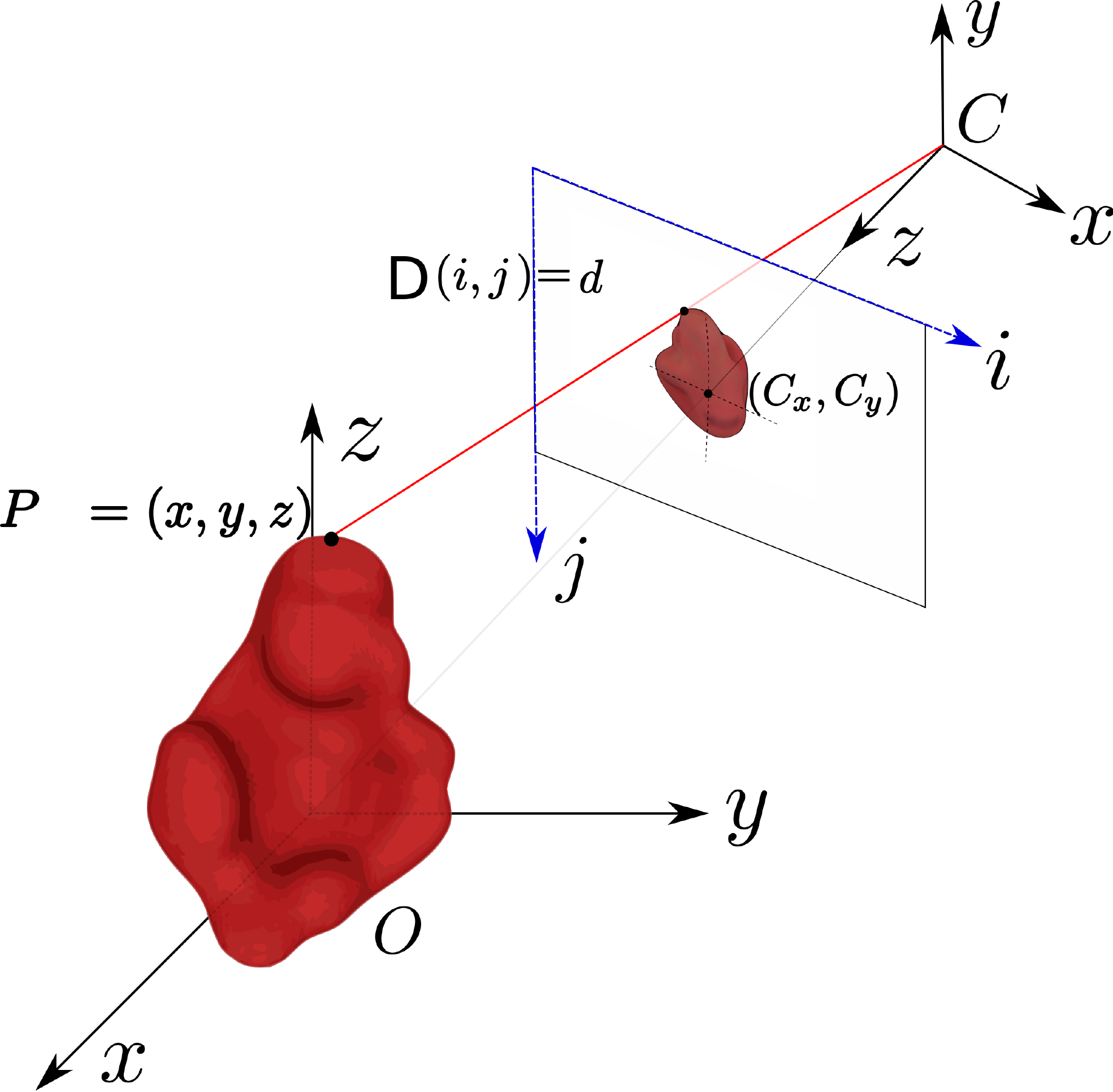}
\caption{Coordinate frames. \textcolor{black}{An $i$ measured point} with respect to the camera's frame is denoted by $P_{i}^{C} = (x^c,y^c,z^c)$, the same point referenced to the global reference frame is denoted $P_{i}^{O} = (x,y,z)$.} \label{fig:marcos_ref}
\end{figure} 
In this step, for all demo objects a discrete set of views (search space) is build. Such views will be the only available poses to place the sensor. Therefore, the perceptions and the NBV are restricted to \textcolor{black}{them}. Formally, the discrete search space is denoted by $V_d$ where $V_d \subset V = \mathbb{R}^{3} \times SO(3)$. To standardize the process, we reuse the method proposed in \cite{hinterstoisser2012model}, where a view sphere around the object and a set of range images are obtained for each view. To be precise, the object is placed in the center of a global reference frame $O$. Then, a set of possible sensor poses, $V_d$, is created by rotating a polar coordinate's point. The magnitude of the polar coordinate's vector is fixed and it provides the distance from the object's center to the sensor position. Next, the sensor poses are obtained by uniformly moving the angles. In our experiments, we use the same discretization provided by \cite{hinterstoisser2012model} (1312 positions). Each view is pointed to the center of the object. Figure \ref{fig:marcos_ref} shows the reference frames and an example of a sensor pose. \\
Once that the views are established, for each demo object we generate the perception that correspond to each view in $V_d$. This step is included for performance reasons because later during training and validation it will be necessary to observe the object from the selected views. Hence, for each view a range image $(D)$ is acquired. Next, the range images are transformed to point clouds, first with respect to the camera's reference frame and later with respect to the global reference frame. At the end of this step, for each view and for each object, we have a point cloud in terms of the global reference frame.
From now on, we will call to each point cloud in the global reference frame a perception and we will write it as $z$. To remark that each perception is generated from a view, we will write a it as $z(v)$. The set of possible perceptions will be written as $Z_d$.

\subsection{Resolution-Optimal Next-Best-View}\label{res_opt_NBV}

The NBV has been ideally defined as the view that increases the amount of information about the object of interest \cite{connolly1985determination}, but in practice, it is usually the view that optimizes a utility function. Such a utility function indirectly measures the increment of information. For example, the functions that count the number of unknown voxels \cite{vasquez2017view} (voxels with probability 0.5) assume a positive correlation between the hidden object's surface and the unknown voxels. The same reasoning applies for the information gain approaches \cite{Isler16}, where it is assumed that reducing the entropy of the partial model will provide more scanned surface. Those approaches are good to provide an approximation of the goodness of the view. However, they do not provide the real NBV because for obvious reasons they do not know the object's shape. Therefore, in this section, we establish criteria for defining the ground truth NBV. The criteria incorporate the fact that the NBV must maximize the scanned surface but also incorporate the overlap that is needed in the real world's reconstructions.

First, let us state some concepts. \textcolor{black}{The set of views where the sensor has been placed is denoted by $S=\{s_0, \dots , s_n\}, S \subseteq V_d$.} The ground truth object is denoted by $W_{obj}$ and it is a point cloud. The object's point cloud should be dense enough, namely, the distance between points should be smaller than the voxel resolution. Also, let us denote the object's accumulated point cloud as $P_{acu}$. Recalling, given that the reconstruction is an iterative integration of perceptions, $P_{acu}$ is the result of integrating the sensor perceptions until the iteration $i$, namely \textcolor{black}{$P_{acu} = \bigcup_{i=0, n} z(s_i)$}. The percentage of coverage of a point cloud $A$ with respect of the object ground truth $W_{obj}$ is computed with the function $\texttt{Coverage}(A, W_{obj})$.

Now, let us define the NBV ($v^*$)  as the view that increases the accumulated point cloud, formally:
\begin{equation}
v^* = \arg\max_{v}  \texttt{Coverage}( z(v) \cup P_{acu},W_{obj})
\end{equation} subject to the following constraints:
\begin{itemize}
\item $v^*$ must be collision free.
\item The perception $z(v^*)$ must have an overlap with $P_{acu}$ higher than a threshold, \textcolor{black}{$\texttt{overlap}(z(v^*),P_{acu})> thresh_1$}. 
\item The common region between surfaces must have at least three 3D features \textcolor{black}{$(thresh_2 \geq 3)$}. Besides of the amount of overlap, 3D features are needed to complete a 3D registration in the three axis. According to \cite{low2007predetermination} at least three features are needed. 
\end{itemize}
In our experiments, we have set $thresh_1$ to 50\% according to the experimental research presented in \cite{vasquez2017view}. To compute the overlap, a radius, called $gap$, is defined to compare the two point clouds, $P_{acu}$ and $z$. Due to the amount of data contained in these point clouds, a KdTree algorithm is needed to make an efficient search over all the neighbors of $P_{acu}$ in $z$ within the radius $gap$. NARF points \cite{steder2010narf} are used as 3D features.

\subsection{Iterative Examples Generation}

To generate the dataset we propose the Algorithm \ref{alg:NBV_dataset}. It reconstructs a given object several times using different initial sensor poses \textcolor{black}{and stores the computed NBVs}. During the reconstruction, the current grid ($M$), accumulated point cloud \textcolor{black}{($P_{acu}$)} and computed NBV \textcolor{black}{($v^*$)} are stored as an example. It is worth to say that, even for the same object a different initial view will produce \textcolor{black}{a different partial model} and a different sequence of views.

The dataset generation requires as input the ground truth object ($W_{obj}$), a correspondent point gap ($gap \leftarrow 0.005$), a stop reconstruction criteria ($S_{cov}$), a maximum number of iterations ($max_{iter}$), an overlap percentage ($thresh_1$), the set of views ($V_d = \{v_1 \dots v_n\}$) and the set of perceptions ($Z = \{ z_1, z_2 \dots z_n \}$). \textcolor{black}{The object is assumed to be reconstructed (stop criterion) when $P_{acu}$ reaches a percentage of coverage equal to $S_{cov}$ or the amount of iterations has reached $max_{iter}$, line 5 of Algorithm \ref{alg:NBV_dataset}. In the experiments, we fixed $max_{iter}$ equal to 10.} Each reconstruction starts from a different initial view $v$, during this process the point cloud $z$ perceived at the pose $v*$ is read and added to the current $P_{acu}$; next, a filtering operation is performed to $P_{acu}$ in order to maintain a uniform density. Then, NARF points are computed in the region created by the overlapped area to assure that the registration process will be performed correctly in the $x$, $y$ and $z$ axes.
Then, in line 9, the probabilistic grid is created or updated, according to the current $P_{acu}$ and $v$. From line 11, an exhaustive search is done looking for the $z$ that maximizes the current $P_{acu}$ compared with $W_{obj}$. To avoid confusions with the current $z$, it is defined a $z'$ to denote a $z$ in evaluation (line 12). In lines 11 to 22 a search of the NBV $v*$, that fulfills the optimal NBV constraints, is done. Finally the algorithm saves a three-tuple which contains $P_{acu}$, $M$ and $v*$. \textcolor{black}{This process is repeated for each $v \in V_d$ as initial view}.
 
\begin{algorithm}[t]
\begin{algorithmic}[1]
	\REQUIRE $W_{obj}$, $gap$, $S_{cov}$, $max_{iter}$, $thresh_1$, $V_d$ and $Z_d$.
	\FOR {$i \leftarrow 1: n$}
	\STATE$v* \leftarrow v_i$ 
	\STATE $iter \leftarrow 0$
	\STATE $P_{acu} \leftarrow \emptyset$
	\STATE $\texttt{Initialize}(M)$
	\WHILE {$\texttt{Coverage}(P_{acu}, W_{obj}) < S_{cov}$ and $iter < max_{iter}$}
	\STATE $P_{acu} \leftarrow  P_{acu} \cup z(v^*) $
	\STATE $P_{acu} \leftarrow \texttt{DownsizeFiltering}(P_{acu})$ 
	\STATE $M \leftarrow \texttt{UpdateGrid}(M, z(v*))$ 
	\STATE $\Delta_{max} \leftarrow 0$
	\FOR {$j\leftarrow 1:n$}
	\STATE $z' \leftarrow \texttt{Perception}(v_j)$
	\IF{$\texttt{overlap}(z', P_{acu}, gap)> thresh_1$}
	\IF {$|\texttt{NARF}(z' \cap P_{acu})| > thresh_2$}
	\STATE {$\Delta \leftarrow \texttt{Coverage}(z' \cup  P_{acu}, W_{obj}) - \texttt{Coverage}(P_{acu}, W_{obj})$}
	\IF {$\Delta > \Delta_{max}$}
	\STATE $v* \leftarrow v_j$
	\STATE $\Delta_{max} \leftarrow \Delta$
	\ENDIF
	\ENDIF
	\ENDIF
	\ENDFOR
	\STATE $iter++$
	\STATE $\texttt{SaveToDataset}(P_{acu}, M, v*)$
	\ENDWHILE
	\ENDFOR
\end{algorithmic}
\caption{Dataset Examples Generation. The algorithm outputs several next-best-views given a known object surface.}\label{alg:NBV_dataset}
\end{algorithm}

\section{NBV Class Prediction using Deep Learning}
Deep learning is a new branch of machine learning. It has improved the classification results provided by the conventional techniques in various scientific domains. The key aspect of deep learning resides in the self-extraction of features or variables learned by the same learning system. Thereby, there is no need for a considerable experience and a careful choice by a user to choose, design and extract a set of features or representations where the performance of the system depends mainly on them. \\ 
Deep learning scheme can be represented by two parts between input data and output data. Composed by the hidden layers or a deep network structure, the first part carries out the features extraction. The second one involves one or two layers which predict the class label. Depending on the problem, the first part can be accomplished by a supervised machine learning approach as the convolutional neural networks, unsupervised one as Auto-Encoders (AEs) or both as the convolutional AEs.\\ 
To face the NBV learning problem, we propose a classification approach, where we consider a possible sensor pose as a class. In the next sections, we will present the discretization proposed to establish the classes and an original 3D convolutional neural network (3D-CNN) architecture to predict the correct class.

\subsection{Classification} \label{section_classif}

In general, a sensor pose is defined in a continuous space. However, to use a classification approach it is necessary to have a discrete set of possible classes. In consequence, we group the poses generating a reduced set. This imply that predictions of the 3D-CNN will be a pose centered in a region representing a group of poses. To create a discrete set of views a sphere tessellation or other techniques can be applied. In our experiments, we create $\mathcal{C} = 14$ sensor poses evenly distributed on the superior half of a sphere. From now on, we will call to each possible pose a class.


\subsection{Architecture}

A 3D convolutional neural network (3D-CNN) is a special case of a CNN. In this kind of architecture, the kernels are 3D volumes instead of 2D maps. In the literature, there are several 3D-CNN architectures for 3D representations, for example VoxNet \cite{maturana2015voxnet} and ShapeNets \cite{wu20153d} are applied to volumes recognition. To the best of our knowledge no 3D-CNN has been applied to the NBV prediction. Consequently, we propose an architecture for NBV class prediction based on the problem's nature.

Henceforth, to simplify the notation, the next shorthand description is used:
\begin{itemize}
\item C(f,k,s): 3D convolution layer with $f$ features, a kernel of size $k\times k\times k$ with stride $s \times s \times s$.
\item P(s): A max pooling operation of stride $s \times s \times s$.
\item FC(n): Fully connected layer with $n$ parameters as output.
\end{itemize}
\begin{figure}[tb]
\centering
\includegraphics[width=\linewidth]{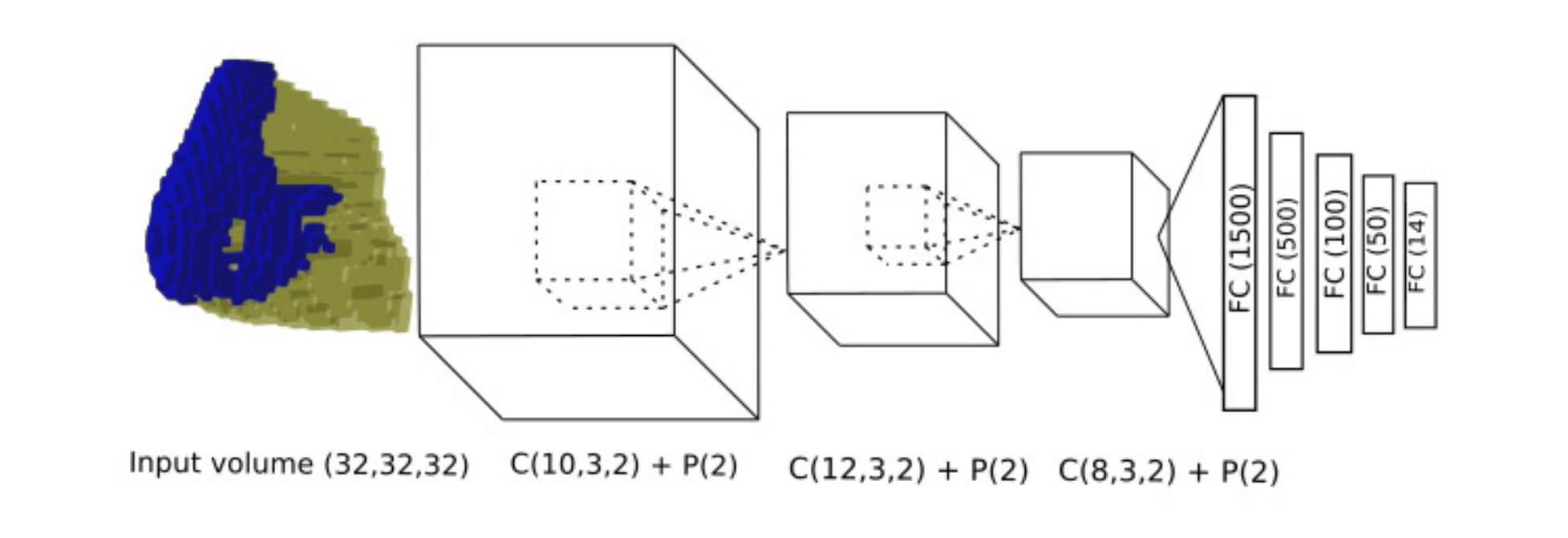}
\caption{\textcolor{black}{NBV-Net architecture.} A volume of size 32 $\times$ 32 $\times$ 32 is fed to the network and 14 possible classes are given as output.} \label{fig:red_3d}
\end{figure}
The proposed architecture, called NBV-Net, is connected as follows: C(10,3,2) $-$ P(2) $-$ C(12,3,2) $-$ P(2) $-$ C(8,3,2) $-$ P(2) $-$ FC(1500) $-$ FC(500) $-$ FC(100) $-$ FC(50) $-$ FC($\mathcal{C} = 14$). After each layer, except polling, the ReLu activation function is used. A softmax function with cross entropy is applied to provide one-hot enconding. Fig. \ref{fig:red_3d} illustrates the network.

\section{Experiments and Results}

\begin{figure*}[tb]
\begin{tabular}{|c|c|c|c|c|c|c|c|c|c|}
	\hline
	\addheight{\includegraphics[width=0.07\textwidth]{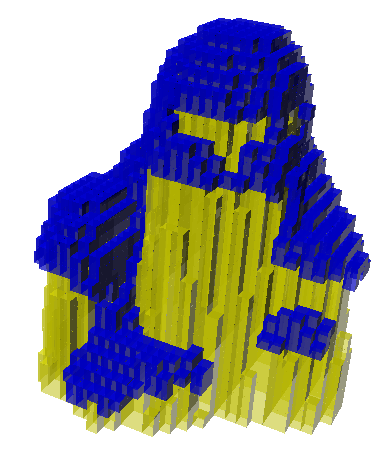}} &
	\addheight{\includegraphics[width=0.07\textwidth]{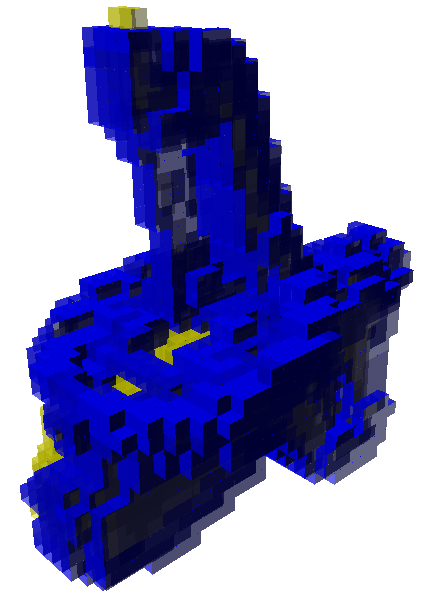}}  &
	\addheight{\includegraphics[width=0.07\textwidth]{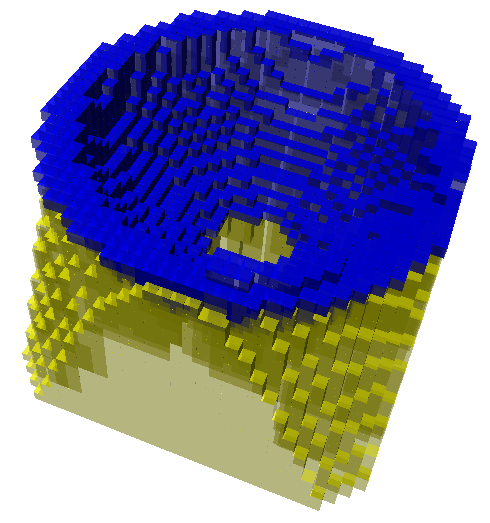}}  &
	\addheight{\includegraphics[width=0.07\textwidth]{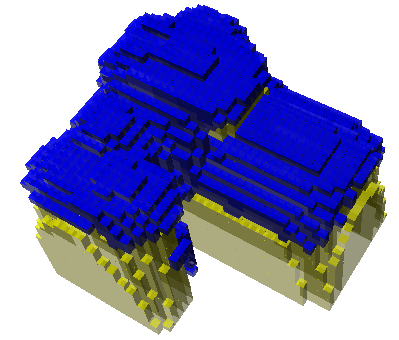}}  &
	\addheight{\includegraphics[width=0.07\textwidth]{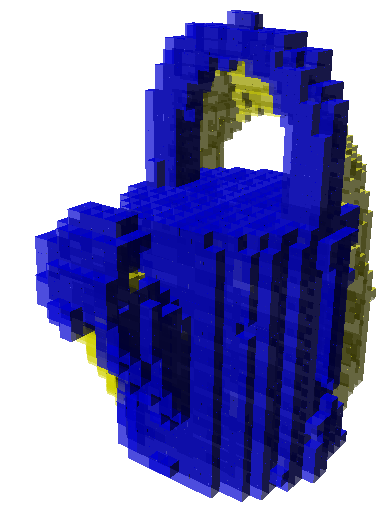}}  &
	\addheight{\includegraphics[width=0.07\textwidth]{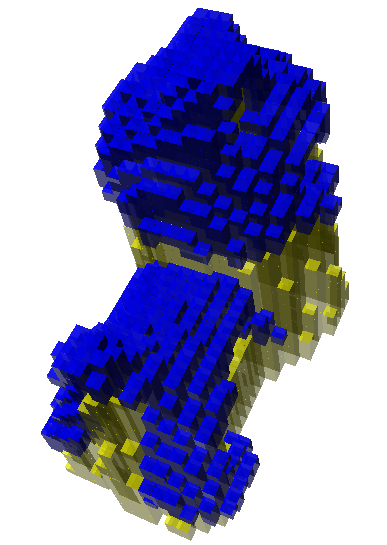}}  &
	\addheight{\includegraphics[width=0.07\textwidth]{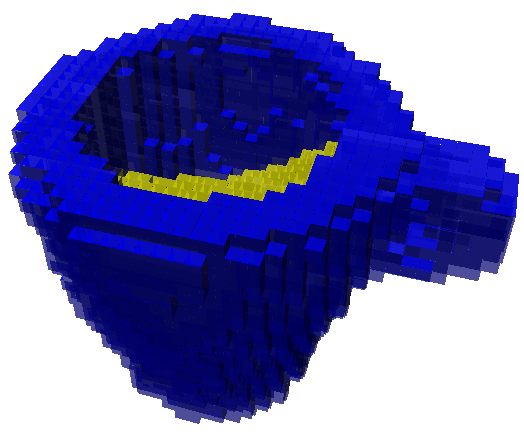}}  &
	\addheight{\includegraphics[width=0.07\textwidth]{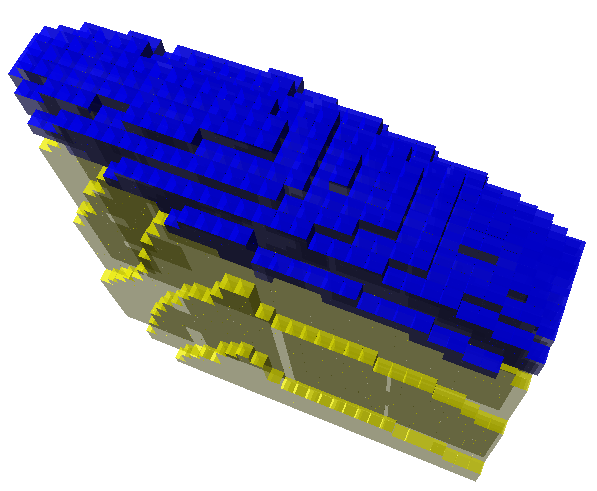}}  &
	\addheight{\includegraphics[width=0.07\textwidth]{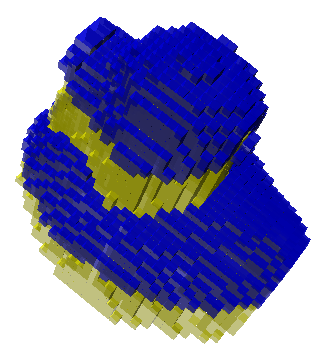}}  &
	\addheight{\includegraphics[width=0.07\textwidth]{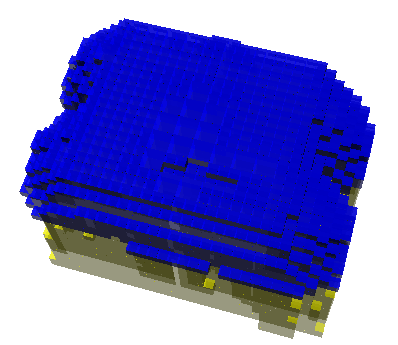}}  \\
	
	\hline
	\addheight{\includegraphics[width=0.07\textwidth]{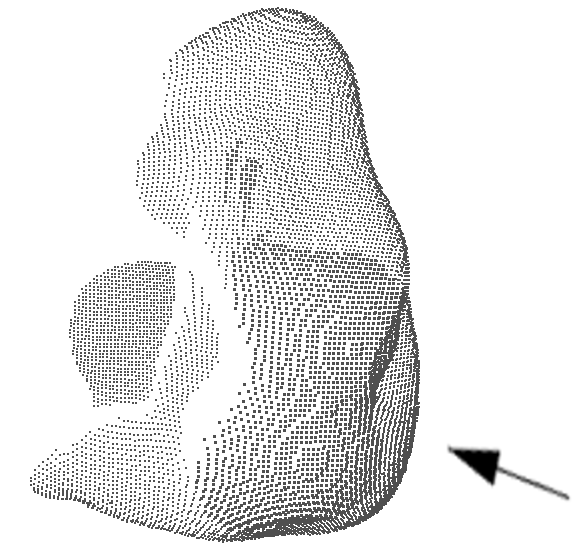}} &
	\addheight{\includegraphics[width=0.07\textwidth]{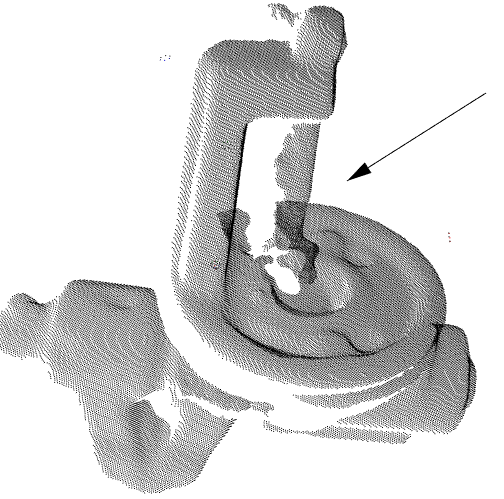}} &
	\addheight{\includegraphics[width=0.07\textwidth]{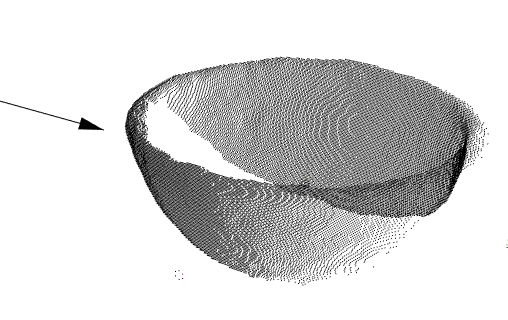}} &
	\addheight{\includegraphics[width=0.07\textwidth]{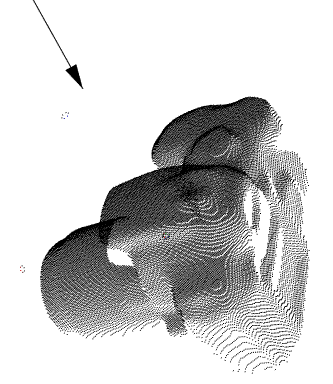}} &
	\addheight{\includegraphics[width=0.07\textwidth]{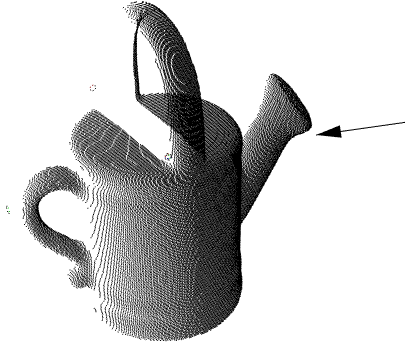}} &
	\addheight{\includegraphics[width=0.07\textwidth]{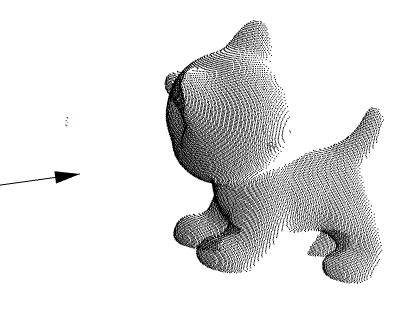}} &
	\addheight{\includegraphics[width=0.07\textwidth]{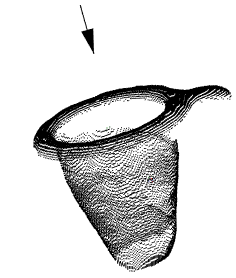}} &
	\addheight{\includegraphics[width=0.07\textwidth]{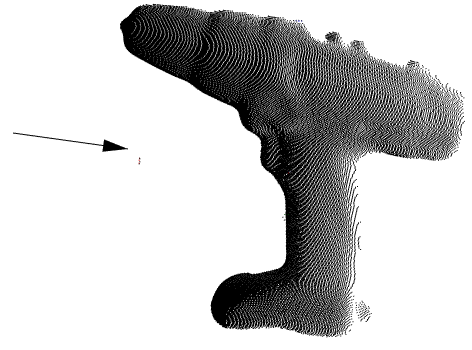}} &
	\addheight{\includegraphics[width=0.07\textwidth]{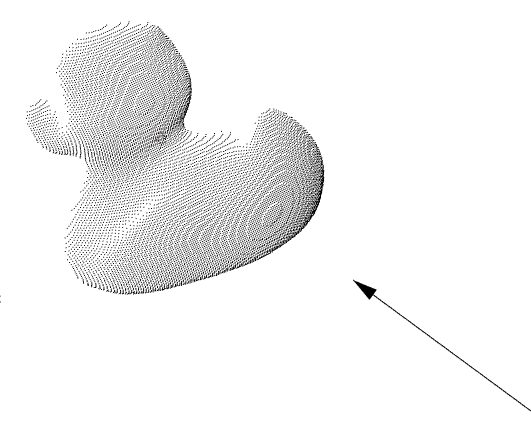}} &
	\addheight{\includegraphics[width=0.07\textwidth]{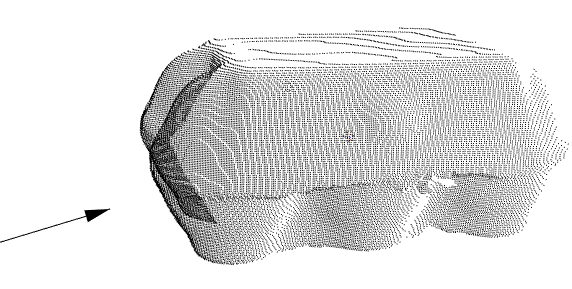}} \\
	\hline
\end{tabular}
\caption{Examples of 10 of 14 objects in the dataset generated for experimentation. Regressors $P_{acu}$ and $M$ with their corresponding responsor $v*$ (NBV). The black arrows show the six parameters which represent the position of a sensor in the space. Blue voxels represent occupied space and yellow voxels the unknown space.}
\label{fig:initital_pose_0}
\end{figure*}

\subsection{Dataset Analysis}
\label{sec:dataset}

The dataset generated for experimentation contains 15, 364 examples that were generated using algorithm \ref{alg:NBV_dataset}. Each example is a tuple of the probabilistic grid and its corresponding NBV. See Fig. \ref{fig:initital_pose_0}. The probabilistic grid has $32 \times 32 \times 32$ voxels and contains the whole object. The NBV designated for each grid is one of the set $V_d$. For this dataset, we created a view sphere with 14 elements. It is worth to say that in \cite{hinterstoisser2012model} the views are only distributed on the superior half of the sphere. Even thought it might appear to be wrong, it is a real world consideration given that objects to be reconstructed are usually placed over a surface, for example a table, therefore, the sensor is unable to see below them. Processing time for generating the complete dataset was about 200 hours. An  Intel$\textregistered$ Core$\texttrademark$ i7-2600 CPU to 3.40GHz with 8GR RAM was employed.

The examples were created using 14 demo objects. Each object was reconstructed from 263 initial views; the views were selected to be uniformly distributed around the view sphere. \textcolor{black}{We did not use the whole $V_d$ set as initial views because we consider that very small differences between initial poses do not provide substantial information.} The stop coverage criterion, $S_{cov}$, was set to $80\%$, this allows a reasonable trade-off between coverage and number of iterations. Distance for correspondent point $gap$ was set to $0.005m$.\\
During the reconstructions, the stop criterion (coverage or a maximum number of iterations) was reached at a different number of iterations \textcolor{black}{due to} the different object shapes and initial view. The objects were reconstructed from 3 to a maximum of 9 iterations and most of them are between 3 and 5 iterations.

\begin{figure}[tb]
\centering
\includegraphics[width=0.5\textwidth, trim=0.5cm 0cm 0.5cm 1cm, clip]{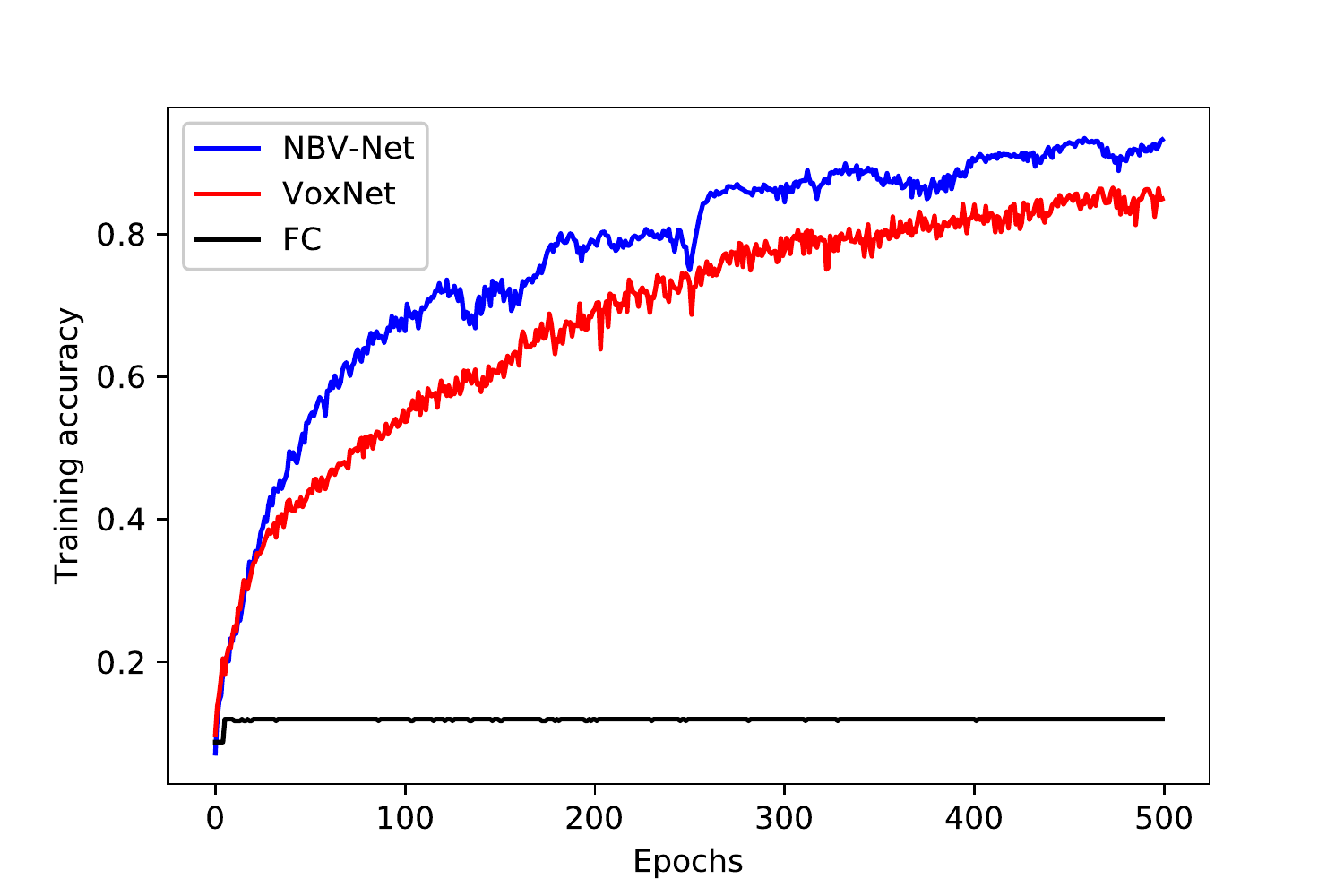}
\caption{\textcolor{black}{Training accuracy for VoxNet, NBV-Net and a fully connected (FC) network.}} \label{fig:error_accur}
\end{figure}

\begin{figure*}[tb]
\subfigure[Bottle]{\includegraphics[scale = 0.4,]{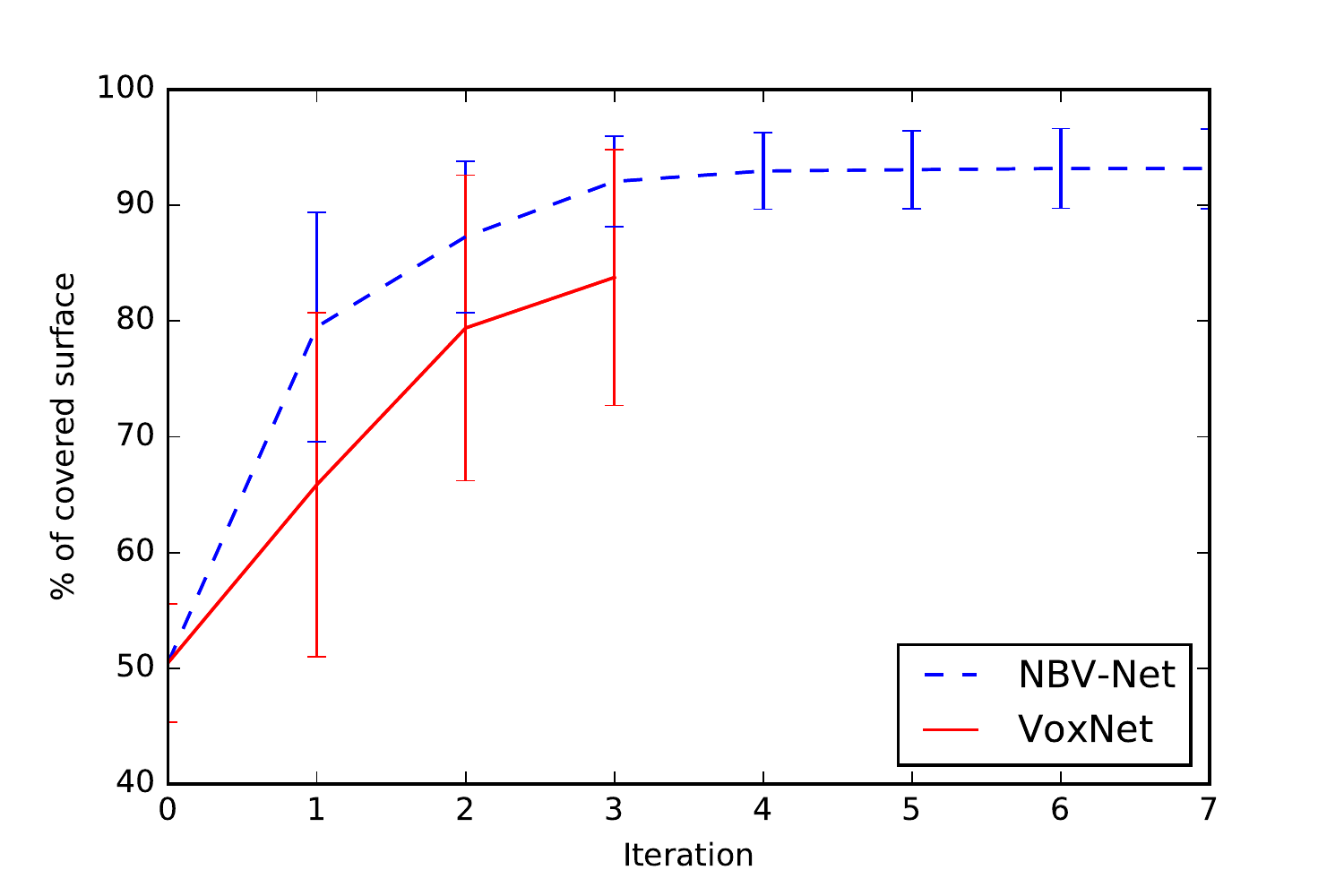}}
\subfigure[Iron]{\includegraphics[scale = 0.4]{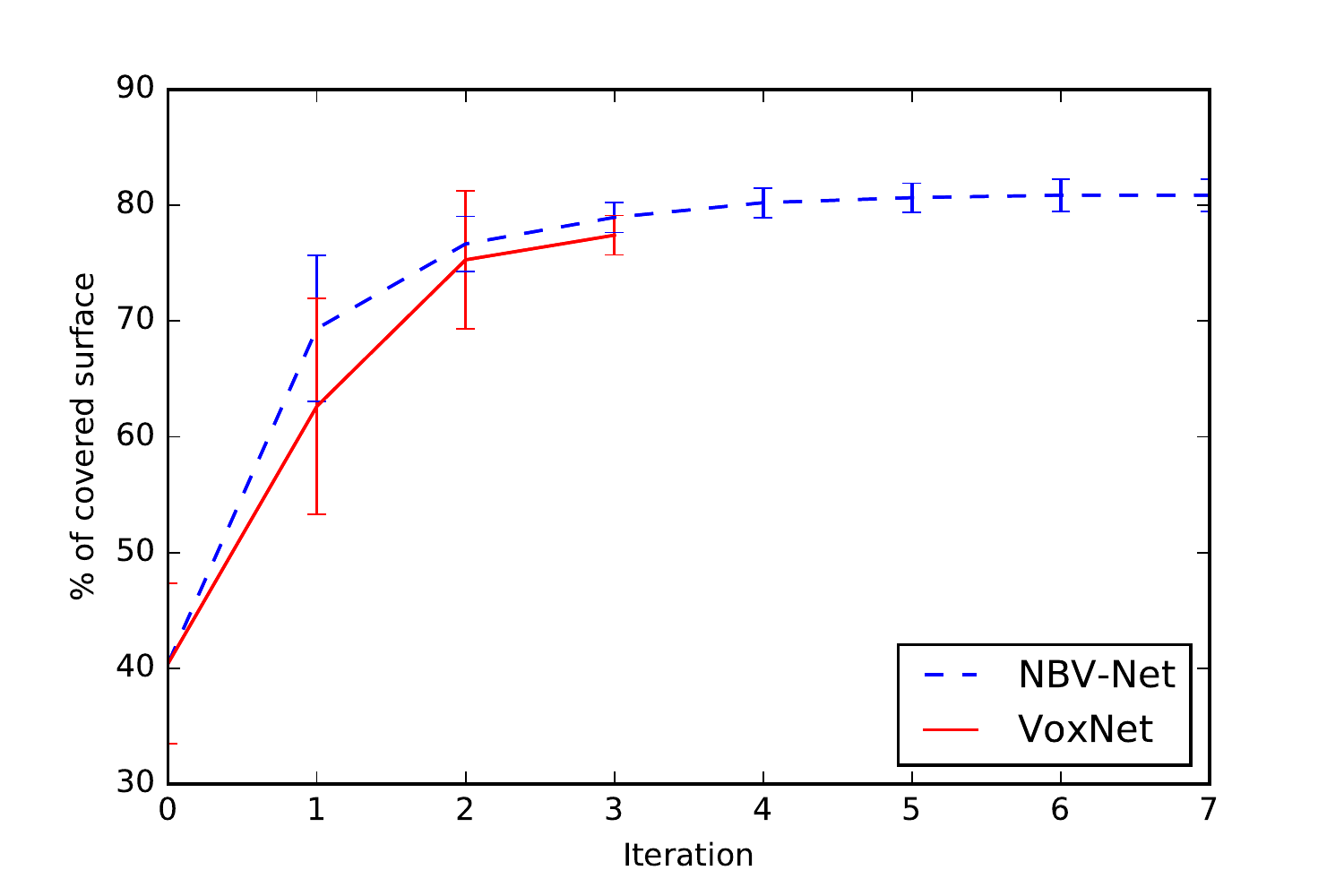}}
\subfigure[Telephone]{\includegraphics[scale = 0.4]{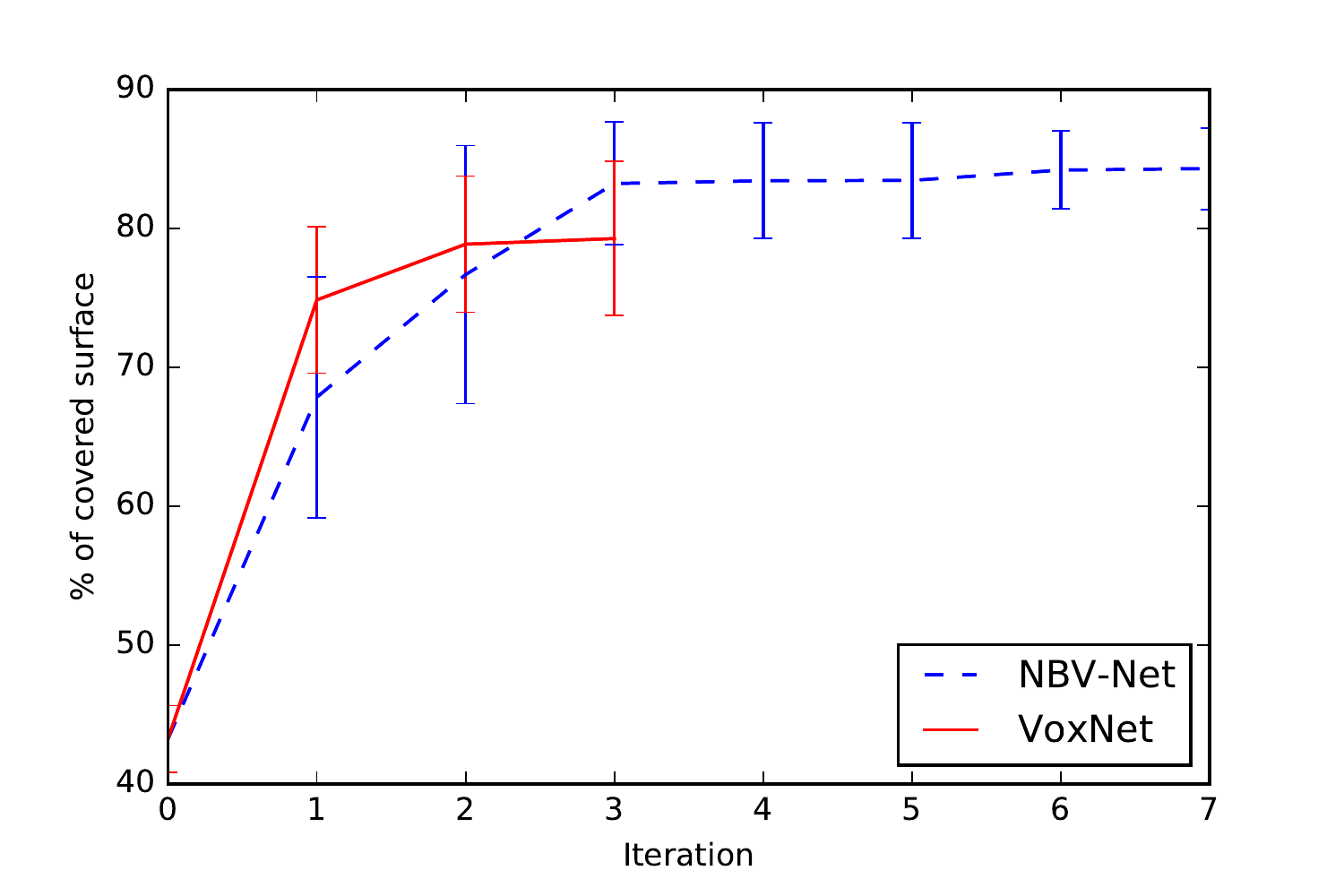}}
\caption{Reconstruction coverage for each object.}
\label{fig:rec_telephone}
\end{figure*}

\subsection{Training}

NBV-Net was trained using the dataset described in section \ref{sec:dataset}. \textcolor{black}{In addition, two supplementary networks were implemented and trained in order to compare the accuracy of NBV-Net.} One of them is a simply fully connected network of four layers; its architecture is FC(1500)-FC(750)-FC(100)-FC(50). We have included the fully connected network as a baseline. The second one is VoxNet \cite{maturana2015voxnet}, even though VoxNet has been designed for object recognition we want to test its performance since its design is similar to the proposed NBV-Net.

\textcolor{black}{Network training was performed by Adam stochastic optimization method \cite{DBLP:journals/corr/KingmaB14}.} The learning rate was 0.001 with a batch size of 200. Dropout, of 0.7, is added after the FC layers and the last convolutional layer. Convolutional layers were initialized with values from a truncated normal distribution with $\sigma = 0.1$ and $\mu = 0$. Our implementation was made in Tensorflow. The training was done using a GPU NVIDIA GeForce GTX1080 mounted on a desktop computer with 16GB of memory and a Xeon E5-2620 v4 2.10GHz processor. Traning for NBV-Net and VoxNet was done using the same hyper-parameters. Training epochs was set to 500. The dataset was splitted in 80\% for training and 20\% for testing.
The training was stopped in 500 epochs, based in the learning of the test set; from this epoch, the test accuracy was oscillating in a margin of $\pm$0.02. Final accuracy reached with each network can be seen in Fig. \ref{fig:error_accur}. It can be seen that the training accuracy of the NBV-Net is better than VoxNet; even when this difference is 0.08, it is important to note that the test set is better learned by the NBV-Net. \textcolor{black}{Processing time for training was 1.2 hours for VoxNet and 1.5 hours for NBV-Net.}

\subsection{3D Reconstruction using Predicted NBV}

\begin{figure}[tb]
\centering
\begin{tabular}{|c|c|c|c|c|c|c|c|}
	\hline
	\small Initial & Iter 1 & Iter 2 & Iter 3 & iter 4  \\
	\hline	
	\addheight{\includegraphics[width = 0.045\textwidth]{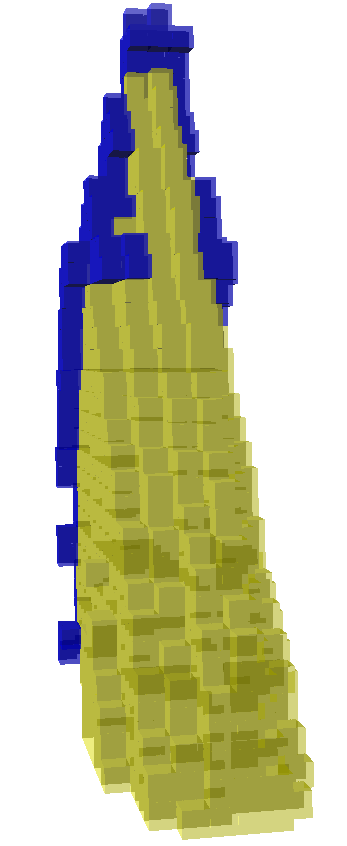}} &
	\addheight{\includegraphics[width = 0.06\textwidth]{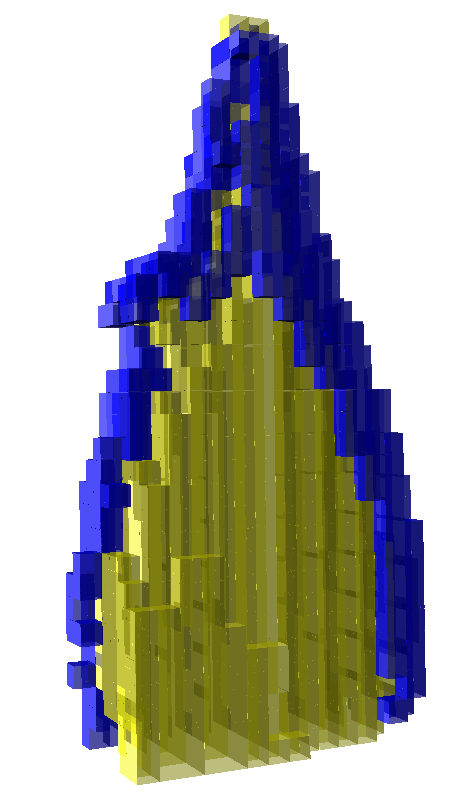}} &
	\addheight{\includegraphics[width = 0.055\textwidth]{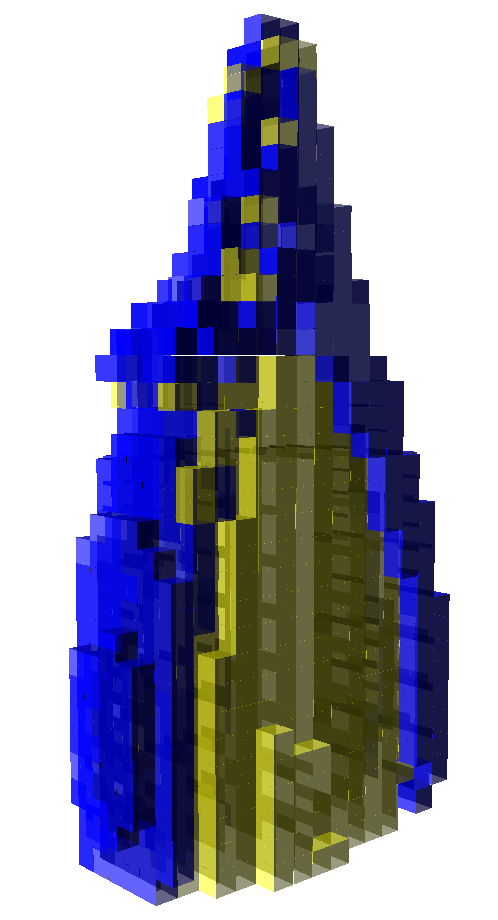}} &
	\addheight{\includegraphics[width = 0.05\textwidth]{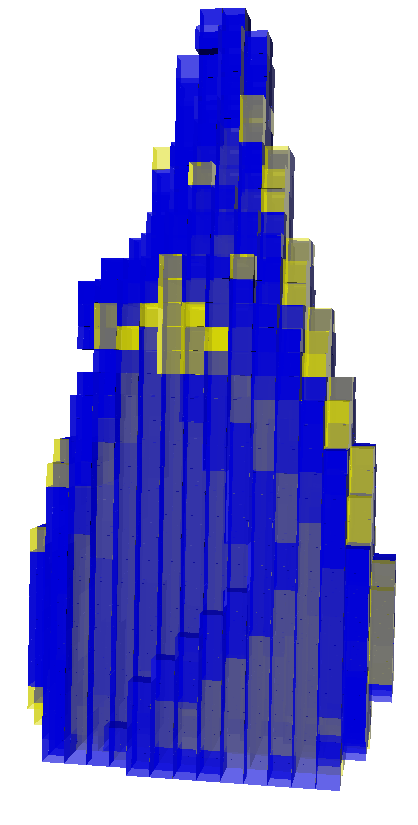}} &
	\addheight{\includegraphics[width = 0.055\textwidth]{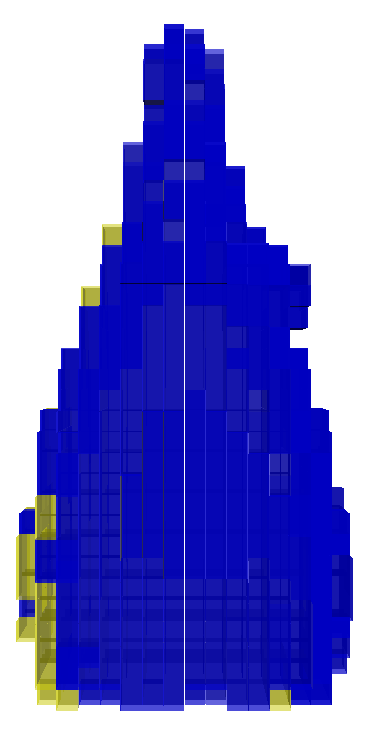}} \\
	55.2\% & 87.1\% & 92.6\% &94.5\%&94.6\%\\
	\hline
	\addheight{\includegraphics[width = 0.08\textwidth]{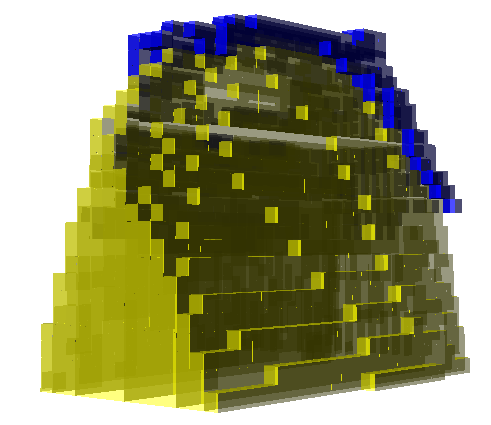}} &
	\addheight{\includegraphics[width = 0.062\textwidth]{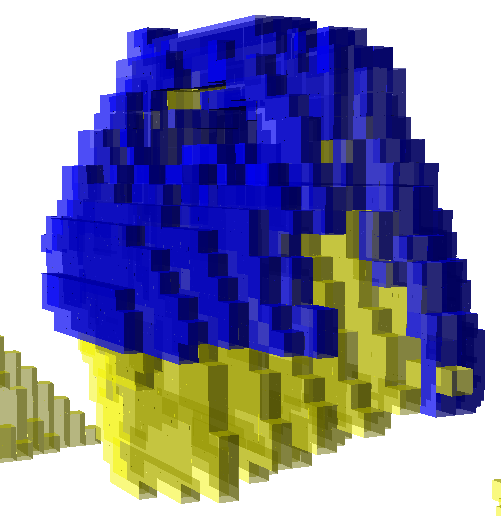}} &
	\addheight{\includegraphics[width = 0.068\textwidth]{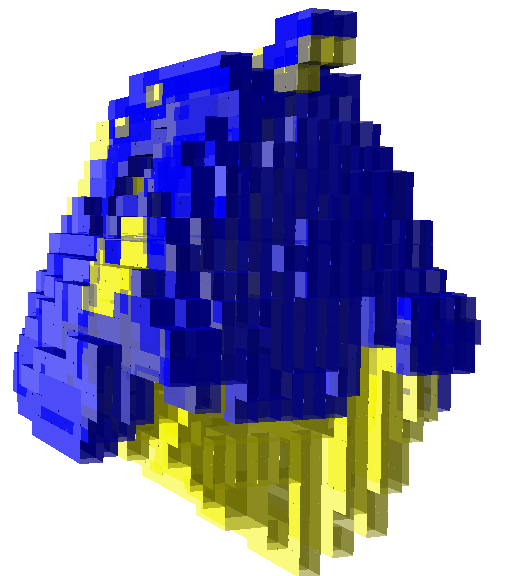}} &
	\addheight{\includegraphics[width = 0.082\textwidth]{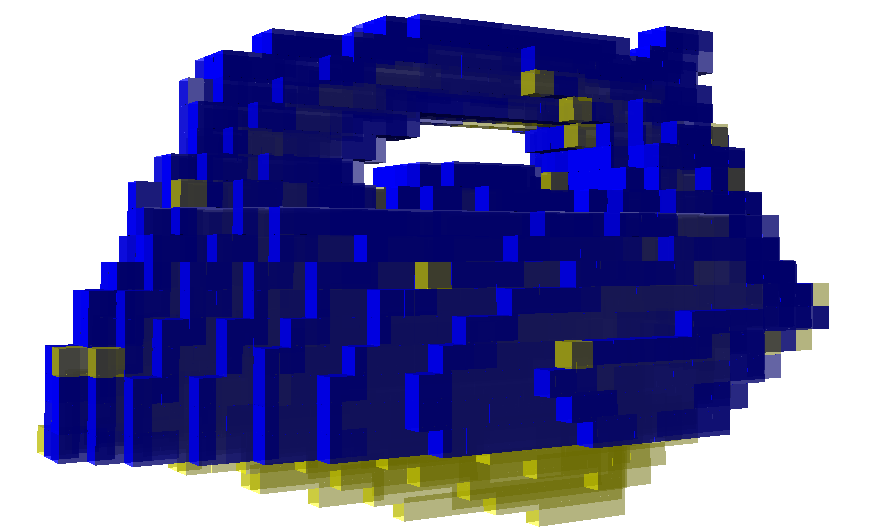}} &
	\addheight{\includegraphics[width = 0.08\textwidth]{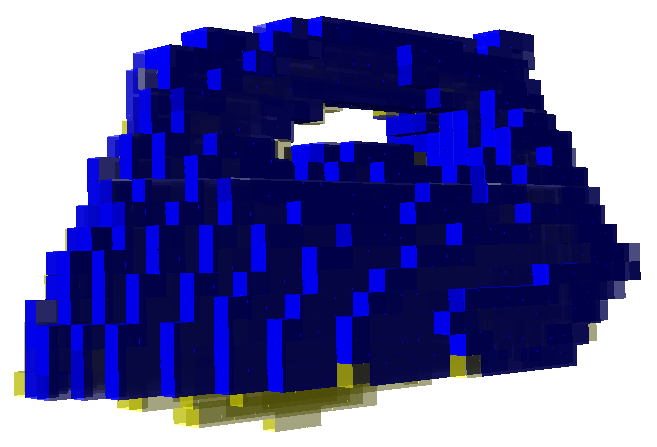}}  \\
	41.5\% & 72.4\% & 77.1\% & 77.7\% & 80.6\% \\
	\hline	
	\addheight{\includegraphics[width = 0.08\textwidth]{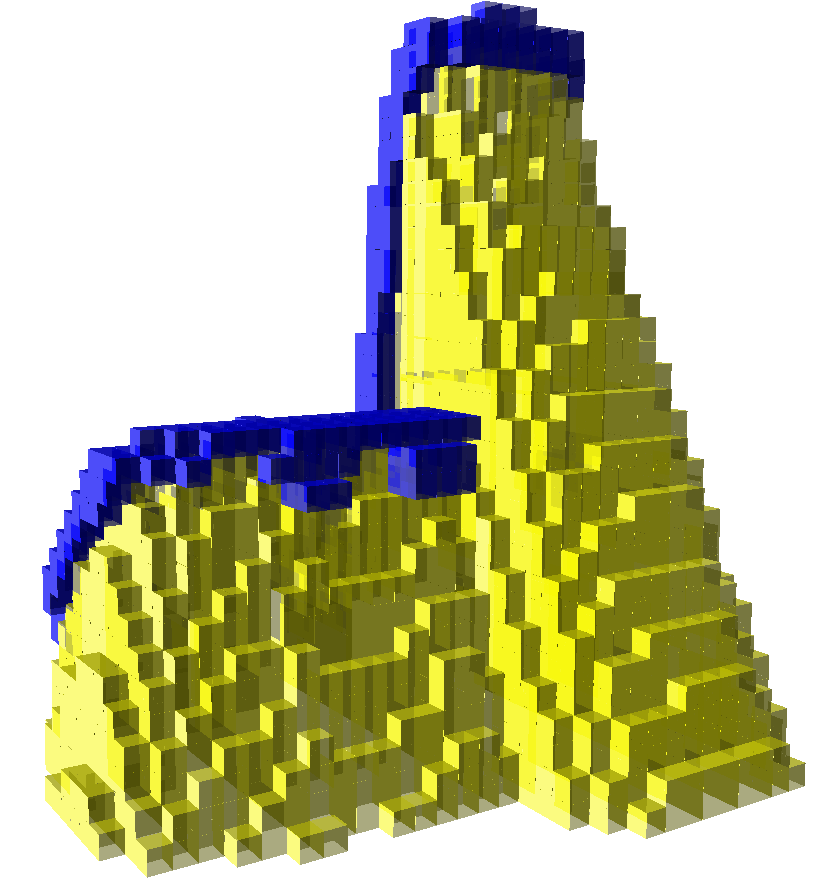}} &
	\addheight{\includegraphics[width = 0.055\textwidth]{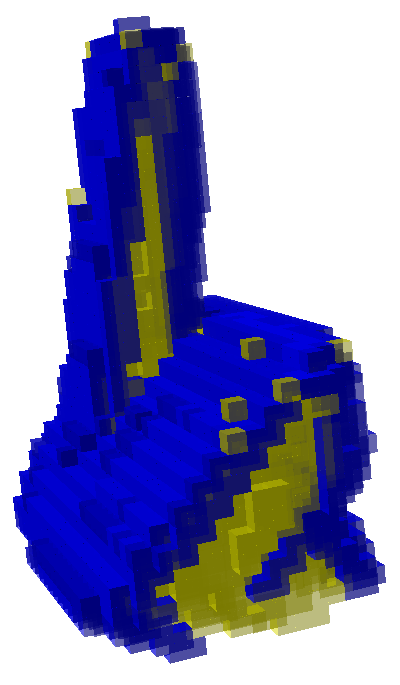}} &
	\addheight{\includegraphics[width = 0.06\textwidth]{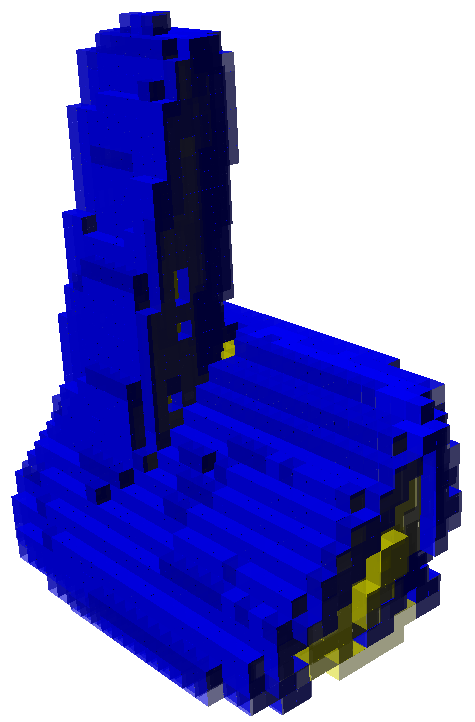}} & &\\
	40.4\% & 77.6\% & 80.6\% & &\\
	\hline
\end{tabular}
\caption{Reconstructed object. In this image the progress of the reconstruction of three objects, using NBV-Net, is shown. Three random initial poses are used for each object. Each frame shows the current state of the probabilistic grid. Blue voxels represent occupied space and yellow voxels the unknown space.}
\label{fig:fig_reconstructed_iters}
\end{figure}

In this experiment, we test the trained networks in a 3D reconstruction task. The objective is to validate that the networks are capable of computing a sensor pose that increases the object's surface. The experiment's workflow is to place the sensor at a random initial pose, then to take a perception, next, the probabilistic grid is updated, then the grid is fed to each trained network and a forward pass is performed, at this moment each network provides a class (the NBV), next, the sensor is placed in the given class and a new perception is taken. The process is repeated until the coverage is not increasing or the network is providing an already visited sensor pose.

The reconstructions were conducted for three new objects: a telephone, an iron, and a bottle. It is important to remark that such objects were not seen by the networks during training. Each object was reconstructed 10 times by each network. The same random initial poses were used for both networks. The results are \textcolor{black}{summarized} in figure \ref{fig:rec_telephone}. The mean and standard deviations of the covered surface are shown in Table \ref{fig:networks_cover}. In \url{https://youtu.be/yw1L4HgDaPI} a video of three reconstructions using NBV-Net is shown.

As a result, NBV-Net reaches a higher coverage in comparison with the other networks. The overlap constraint (Section \ref{res_opt_NBV}) is more evident in NBV-Net because the increments in percentage of covered surface are softer than VoxNet. In Fig. \ref{fig:fig_reconstructed_iters}, the three objects were reconstructed, starting from three random poses and the iterations of the NBV predicted using NBV-Net are shown. \textcolor{black}{The processing time to perform a network forward pass is 1.9s.}

\begin{table}[]
\centering
\caption{The average coverage ($S$), standard deviation ($\sigma_s$), and iterations ($I$) are shown as result of 10 reconstructions from random initial poses.}
\begin{tabular}{|l|l|l|l|l|l|l|}
	\hline
	\multirow{2}{*}{Object} & \multicolumn{3}{c|}{NBV-Net}         & \multicolumn{3}{c|}{VoxNet}          \\ \cline{2-7} 
	& $S_{cov}$ & $\sigma_s$ & $I$ & $S_{cov}$ & $\sigma_s$ & $I$ \\ \hline
	Bottle                    & 93.18           & 3.44       & 4.1     & 83.88           & 11.05      & 2.2     \\ \hline
	Iron                    & 80.77           & 1.41       & 4.5     & 77.64           & 1.8        & 2.6     \\ \hline
	Telephone                    & 84.07           & 3.12       & 3.5     & 79.52           & 5.65       & 1.8     \\ \hline
\end{tabular}
\label{fig:networks_cover}
\end{table}

\section{Conclusions}

A scheme for computing the NBV for 3D object reconstruction based on supervised deep learning has been proposed. As part of the scheme, we have presented an algorithm for automatic generation of datasets and an original 3D convolutional neural network called NBV-Net. We have trained the network and we have compared its accuracy against an alternative net. \textcolor{black}{We also have tested NBV-Net in the reconstruction of several unknown objects which were not seen during training. As a result, NBV-Net was capable of predicting the NBV covering the majority of the surface.} In addition, the computation time for computing the NBV is very short, competing with state-of-the-art NBV methods. In consequence, we have shown positive evidence that the NBV problem can be addressed as a classification problem when supervised deep learning is applied. Our future research interest is to face the NBV learning problem as a regression problem avoiding the restriction of a limited set of classes.


\bibliographystyle{plain}
\bibliography{bibliography}

\end{document}